\documentclass{article}
\usepackage{spconf,amsmath,graphicx}

\let\oldbibliography\thebibliography
\renewcommand{\thebibliography}[1]{%
  \oldbibliography{#1}%
  \setlength{\itemsep}{0pt}%
}

\usepackage{nicefrac,amssymb}
\usepackage{multirow}
\usepackage{tikz}
\usepackage{pgfplots}
\usepackage{caption}
\usepackage{subcaption}
\usepackage{hyperref}

\usepackage{adjustbox}

\usepackage{tikz}
\usepackage{pgfplots}
\usepackage{caption}
\usepackage{subcaption}
\usepgfplotslibrary{colorbrewer}
\pgfplotsset{cycle list/Set1-9}
\tikzset{every picture/.style={line width=1pt}}
\usetikzlibrary{patterns}
\usepgfplotslibrary{fillbetween}
\usepackage[eulergreek]{sansmath}
\usepackage{wrapfig}
\usepackage{tcolorbox}
\usepackage{etoolbox}
\usepackage{rotating}

\usepackage{cite}
\usepackage[overlay,absolute]{textpos}

\title{A Consensual Collaborative learning method for remote sensing image classification under noisy multi-labels}

\name{Ahmet Kerem Aksoy, Mahdyar Ravanbakhsh, Tristan Kreuziger, Beg{\"u}m Demir}
\address{Faculty of Electrical Engineering and Computer Science, Technische Universit{\"a}t Berlin, Germany}
\begin{document}
\ninept
\begin{textblock*}{10in}(5mm, 5mm)
{\scriptsize{$\copyright$ 2021 IEEE. Published in the IEEE 2021 IEEE International Conference on Image Processing (ICIP 2021), September 2021, Anchorage, Alaska, USA. Personal use of this material is permitted.

}}
\end{textblock*}
\maketitle
\begin{abstract}
 Collecting a large number of reliable training images annotated by multiple land-cover class labels in the framework of multi-label classification is time-consuming and costly in remote sensing (RS). To address this problem, publicly available thematic products are often used for annotating RS images with zero-labeling-cost. However, such an approach may result in constructing a training set with noisy multi-labels, distorting the learning process. To address this problem, we propose a Consensual Collaborative Multi-Label Learning (CCML) method. The proposed CCML identifies, ranks and corrects training images with noisy multi-labels through four main modules: 1) discrepancy module; 2) group lasso module; 3) flipping module; and 4) swap module. The discrepancy module ensures that the two networks learn diverse features, while obtaining the same predictions. The group lasso module detects the potentially noisy labels by estimating the label uncertainty based on the aggregation of two collaborative networks. The flipping module corrects the identified noisy labels, whereas the swap module exchanges the ranking information between the two networks. The experimental results confirm the success of the proposed CCML under high (synthetically added) multi-label noise rates. The code of the proposed method is publicly available at \href{https://noisy-labels-in-rs.org}{https://noisy-labels-in-rs.org}.

\end{abstract}
\section{Introduction}
\label{sec:intro}
Remote sensing (RS) images acquired by satellite or airborne sensors are a rich source of information for Earth observation (EO) \cite{bruzzone2014review}. One of the growing research interests in RS is to develop multi-label RS image scene classification methods that aim to automatically assign multiple land-cover class labels (i.e., multi-labels) to each image. In recent years, deep learning (DL) based methods have attracted great attention in multi-label classification (MLC) of RS images \cite{SigmoidCNN, Zeggada:2017, 8451836, Alshehri:2019, RsimCNN,yessou2020comparative}. Most of the existing methods require a high number of reliable multi-label RS images for training such DL models, which is time-consuming, costly and needs expertise to gather. For constructing a large training set with zero-annotation effort, publicly available thematic products (e.g., the Corine Land Cover [CLC] map, the GLC2000 and the GlobCover) that are available in RS can be used as labeling sources \cite{buttner2004corine, paris2020unsuperextract}. However, the set of land-cover class labels provided through the thematic products can be noisy (i.e., incomplete or wrong). Using training images with noisy labels may result in uncertainty in the MLC model, leading to reduced performance in multi-label predictions. Therefore, there is a need to develop methods that are robust to uncertainty in class labels.


In general, the sources of uncertainty can be characterized into two categories \cite{collier2020simple}: i) aleatoric uncertainty; and ii) epistemic uncertainty. 
Aleatoric uncertainty is associated to the inherent noise in the input data (e.g., label noise), and can be homoscedastic (which is associated to constant uncertainty across the training data) or heteroscedastic (when the uncertainty varies across the training data). The epistemic uncertainty is associated to uncertainty over the parameters of the considered models. It is shown that the epistemic uncertainty decreases when the amount of input data significantly increases, while aleatoric uncertainty always persists \cite{collier2020simple}. In this paper, our focus is to estimate aleatoric uncertainty for MLC problems in the presence of multi-label noise. In general, for RS images annotated by multi-labels, two types of label noise can be present in a given image: 1) noise associated to missing labels (occurs when a land-cover class label is not assigned to the image although it is present in the image) \cite{jain2016extreme}; 
2) noise associated to wrong labels (occurs when a land-cover class label is assigned to the image although it is not present in the image) \cite{durand2019learning}. 

According to our knowledge, there are only few studies that aim to overcome the negative effect of noisy labels in MLC problems in RS. As an example, Ulmas et al. introduce a semantic segmentation method that identifies label noise by evaluating the loss values, considering that noisy labels are associated with the highest losses \cite{ulmas2020segmentation}. As another example, Hua et al. present a regularization method, which is defined based on a label correlation matrix constructed by the semantic word embedding of the labels \cite{igarssnoisy2020}. The underlying assumption is that semantically similar classes are likely to be present together. Both methods can only identify the wrong label noise, while ignoring the missing label noise.

In this paper, we propose a novel Consensual Collaborative Multi-label Learning (CCML) method. The proposed CCML aims at estimating aleatoric uncertainty on label annotations without making any prior assumption about the type of noise in multi-label training sets. To this end, the proposed CCML method includes four modules: 1) discrepancy module; 2) group lasso module; 3) flipping module; and 4) swap module. The discrepancy module aims at forcing the two networks to learn diverse features, while achieving consistent predictions. The group lasso module is proposed to identify the type of label noise automatically. To identify images with noisy labels, the group lasso module computes a sample-wise ranking loss and estimates aleatoric uncertainty. The flipping module flips the labels associated with higher uncertainty. The tasks of the swap module are first to exchange the ranking information between the two networks and then to exclude the images with extremely noisy labels from the back-propagation. 

The main contributions of this paper can be summarized as follows: 
i) to the best of our knowledge, the CCML is the first method that can identify the two types of multi-label noise without any prior assumption;
ii) CCML is robust against high label noise rates in the training sets; and iii) CCML is an architecture-independent method, which can be combined with different classification approaches to detect the potentially noisy labels assigned to the training images with multi-labels.


\section{proposed method}
\label{sec:method}
Let $\mathbf{D}=\{\mathbf{x}_{1},\mathbf{x}_{2},\dots,\mathbf{x}_{M}\}$ be an archive that consists of $M$ images, where $\mathbf{x}_{i}$ indicates the $\mathit{i}^{th}$ image in the archive. Each image in the archive is associated with one or more classes from a set of labels $\{l_1,l_2,\dots,l_V\}$ where $V$ is the total number of classes. Let $\mathbf{y}_i=[{y}_{i}^1,{y}_{i}^2,\ldots,{y}_i^V]$ be a binary vector, where ${y}_i^j$ indicates the presence or absence of a label $l_j$ for the image $\mathbf{x}_i$. Thus, the multi-labels of the $i$-{th} image $\mathbf{x}_i$ are given by $\mathbf{y}_i$. We assume that the labels $\mathbf{y}_{i}$ can be noisy (e.g., wrong or missing). To reduce the negative effect of the noisy labels during the training process, we propose a novel Consensual Collaborative Multi-label Learning (CCML) method. CCML is defined based on two identical CNNs $\mathit{f}$ and $\mathit{g}$ with parameters $\boldsymbol\theta$ and $\boldsymbol{\hat{\theta}}$, respectively. For each network, the Binary Cross Entropy (BCE) is chosen for the classification loss function as suggested in \cite{sumbul2019bigearthnet}. The considered collaborative networks aim at: i) learning diverse features, while predicting the same class distribution; and ii) being capable of correcting errors of each other during the course of training through information exchange. To this end, the proposed CCML contains four modules: 1) discrepancy module; 2) group lasso module; 3) flipping module; and 4) swap module. These modules are described in the following.

\begin{figure}[t]
  \centering
  \includegraphics[width=\linewidth]{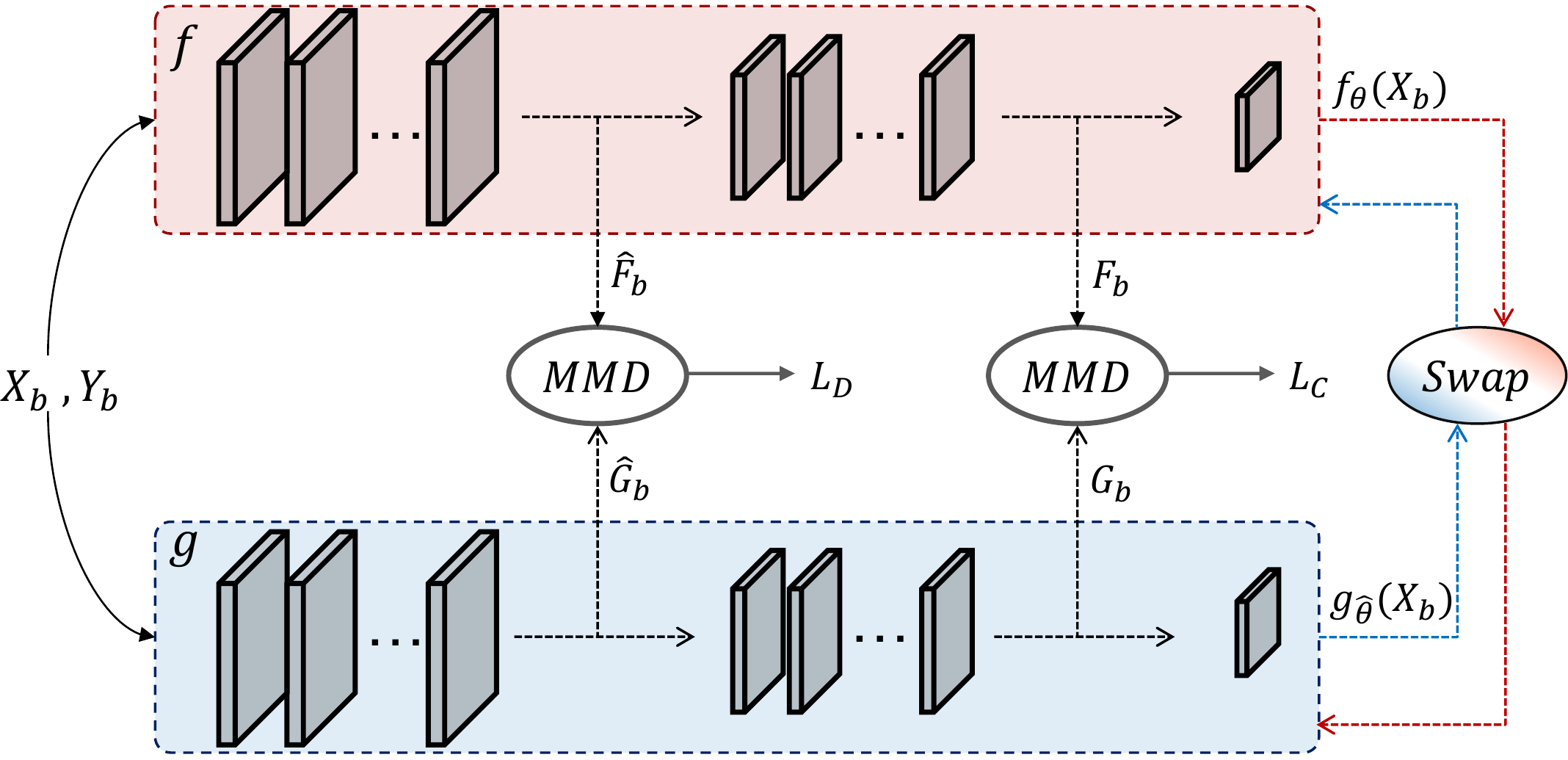}
  \captionsetup{width=\linewidth}
  \caption{Block diagram of the training phase of the proposed CCML. $\mathit{f}$ and $\mathit{g}$ represent the two collaborative networks with parameters $\boldsymbol\theta$ and $\boldsymbol{\hat{\theta}}$, respectively. $\mathbf{\hat F}_{b}$ and $\mathbf{\hat G}_{b}$ denote the intermediate logits of the networks. $\mathbf{F}_{b}$ and $\mathbf{G}_{b}$ denote the logits of the last layers. \emph{MMD} represents the discrepancy modules, and \emph{Swap} is the swapping module for exchanging the ranking information.}
  \label{fig:mmd}
\end{figure}

\noindent\textbf{Discrepancy Module:}
This module aims at forcing the two collaborative networks to learn different features, while ensuring consistent predictions. To this end, we consider two discrepancy modules for calculating two loss functions: i) the disparity loss ($\mathit{L_{D}}$); and ii) the consistency loss ($\mathit{L_{C}}$). The disparity loss function ensures that the networks learn distinct features, while the consistency loss function ensures that the two networks produce similar predictions. Therefore, the disparity loss is calculated between the chosen layers of the networks, and the consistency loss is calculated at the end of the networks (See Fig. \ref{fig:mmd}). The final loss functions $\mathit{Loss}_{f}^{b}$ and $\mathit{Loss}_{g}^{b}$ to be minimized for $f$ and $g$ are defined as:
\begin{equation} \label{eq:3.3}
\begin{split}
\mathit{Loss}_{f}^{b} = \frac{1}{R}\sum_{i=1}^{R} \mathit{L}_{f} (\mathbf{x}_{i}) + \lambda_{1}\mathit{L_{C}} - \lambda_{2}\mathit{L_{D}}  \; \;
\\
\mathit{Loss}_{g}^{b} = \frac{1}{R}\sum_{i=1}^{R} \mathit{L}_{g} (\mathbf{x}_{i}) + \lambda_{1}\mathit{L_{C}} - \lambda_{2}\mathit{L_{D}}  \; \; ,
\end{split}
\end{equation}
\noindent where $\mathit{L}_{f}$ and $\mathit{L}_{g}$ denote the BCE loss functions for the networks $f$ and $g$, respectively. $\mathit{R}$ represents the number of selected samples associated to small loss values from the mini-batch $\mathit{b}$. $\lambda_{1}$ and $\lambda_{2}$ represent the combination weights for the consistency loss $\mathit{L_{C}}$ and disparity loss $\mathit{L_{D}}$, respectively. The disparity loss $\mathit{L_{D}}$ is defined as:
\begin{equation} \label{eq:3.5}
\begin{split}
\mathit{L_{D}} = \mathit{MMD}(\mathbf{\hat{F}}_{b},\mathbf{\hat{G}}_{b}) \; ,\\
\mathbf{\hat{F}}_{b} = \mathit{f_{\theta(1:l)}}(\mathbf{X}_{b})\;
,
\mathbf{\hat{G}}_{b} = \mathit{g_{\hat{\theta}(1:l)}}(\mathbf{X}_{b})\; ,\\
\end{split}
\end{equation}
\noindent where $\mathit{MMD}$ is the discrepancy module. $\mathbf{\hat{F}}_{b}$ and $\mathbf{\hat{G}}_{b}$ represent the logits of the layers before the module.  The parameters of the networks $f$ and $g$ up to the layer $\mathit{l}$ denoted as $\boldsymbol\theta(1:l)$ and $\boldsymbol{\hat{\theta}}(1:l)$, respectively. $\mathit{l}$ is the layer that the discrepancy module for the disparity loss is inserted. The set of training images in the given mini-batch $b$ is denoted as $\mathbf{X}_{b}$. The consistency loss $\mathit{L_{C}}$ is defined as: 
  
\begin{equation} \label{eq:3.6}
\mathit{L_{C}} = \mathit{MMD}(\mathbf{F}_{b},\mathbf{G}_{b}) \; ,
\end{equation}

\noindent
where $\mathbf{F}_{b}$ and $\mathbf{G}_{b}$ denote the logits of the last layer of the networks, analogous to (\ref{eq:3.5}). The discrepancy module is a statistical distance function, which measures the difference between two probability distributions. We select the Maximum Mean Discrepancy (MMD)\cite{gretton2012kernel} algorithm to be used in the discrepancy module due to its success to disentangle probability distributions \cite{han2020learning}. The MMD for two distributions $\mathit{P}$ and $\mathit{Q}$ can be defined as:  
\begin{equation} \label{eq:3.7}
\mathit{MMD(P,Q)} = \mathit{\| \boldsymbol{\mu}_{P} - \boldsymbol{\mu}_{Q} \|_{H}}  \; \; ,
\end{equation}     

\noindent
where $\boldsymbol{\mu}_{P}$ and $\boldsymbol{\mu}_{Q}$ denote the mean values of the distributions $\mathit{P}$ and $\mathit{Q}$, respectively. $\mathit{H}$ denotes the Reproducing Kernel Hilbert Space (RKHS)\cite{alvarez2011kernels} and $\| \|_{H}$ represents the $L_1$ norm. An empirical estimation of MMD between $P$ and $Q$ can be denoted as:  
\begin{equation} \label{eq:3.8}
\begin{split}
\mathit{MMD(P,Q)} = \frac{1}{m^{2}} \bigg[ \sum_{i=1}^{m} \sum_{t=1}^{m} \mathit{k(\mathbf{s}_{i}^{P},\mathbf{s}_{t}^{P})} -\\ 2 \sum_{i=1}^{m} \sum_{t=1}^{m} \mathit{k(\mathbf{s}_{i}^{P},\mathbf{s}_{t}^{Q})} 
+ \sum_{i=1}^{m} \sum_{t=1}^{m} \mathit{k(\mathbf{s}_{i}^{Q},\mathbf{s}_{t}^{Q})} \bigg]  \; \; ,
\end{split}
\end{equation}

\noindent  where $\mathbf{s}_{i}^{P}$ and $\mathbf{s}_{i}^{Q}$ are $i$-th samples from respective distributions. $k$ is the Gaussian radial basis function kernel \cite{vert2004primer}.

\noindent\textbf{Group Lasso Module:}
Although the BCE loss function is commonly used for the multi-label classification tasks, it is not robust to label noise in the training sets and may perform inadequately as shown in \cite{han2020learning}. Therefore, there is a need for an additional mechanism to avoid the model deceived by noisy training sets. Inspired by \cite{bucak2011multi}, we propose a ranking error function to estimate uncertainty on label annotations. Such a function can deal with both types of noise in a multi-label scenario without making any prior assumption. The motivation behind using the group lasso is to estimate aleatoric uncertainty using the predictions of the two networks to identify potentially noisy labels in the training set. Furthermore, the group lasso provides information about label noise types (i.e., missing and wrong labels). Let $\mathit{E_{c,\hat c}}$ denote the ranking error function:
\begin{equation}
\mathit{E_{c,\hat c}(\mathbf{x}_{i})} = \max\Big(0  \;, \; 2\big[f_{\hat c}(\mathbf{x}_{i})-f_{c}(\mathbf{x}_{i})\big]+ 1 \Big) ,
\end{equation}
\noindent where $\mathit{c}$ represents a class index associated by value $1$ for assigned labels and $\mathit{\hat c}$ represents a class index associated by 0 (which represents labels that are not assigned) for image $\mathbf{x}_{i}$ in the given batch. $f_{c}(\mathbf{x}_{i})$ and $f_{\hat c}(\mathbf{x}_{i})$ denote the predictions of the classes $\mathit{c}$ and $\mathit{\hat c}$, respectively. The ranking error function gives a measure of the potential noise in a class combination. In the case of an accurate prediction the ranking error is equal to $0$. Otherwise the ranking error function returns a positive value indicating label noise. The main reason is that if $\mathit{c} \in \mathbf{y}_{i}$ and $\mathit{\hat c} \notin \mathbf{y}_{i}$, then $f_{c}(\mathbf{x}_{i}) > f_{\hat c}(\mathbf{x}_{i})$. The ranking loss is defined as:
\begin{equation} \label{eq:3.10}
\mathit{Lasso_{f}(x_{i})} = \alpha \sum_{\hat c=a+1}^{m} \sqrt{ \sum_{c=1}^{a} \epsilon } + \beta  \sum_{c=b+1}^{m} \sqrt{ \sum_{\hat c=1}^{b} \epsilon } \;,
\end{equation}
\noindent
where $\epsilon = E^{2}_{c,\hat c}(\mathbf{x}_{i})$ is the ranking error, while $\mathit{a}$ and $\mathit{b}$ denote the assigned and unassigned labels to the image $\mathbf{x}_{i}$, respectively. The first loss term calculates an aggregated loss based on missing class labels, while the second term calculates an aggregated loss for wrong class label assignments. This approach allows our method to rank training samples associated with noisy labels according to their estimated noise rate and noise type by adjusting the importance factors ($\alpha$ and $\beta$) of the loss terms. $\mathit{Lasso_{g}(\mathbf{x}_{i})}$ is computed analogous to $\mathit{Lasso_{f}(\mathbf{x}_{i})}$. The returned ranking losses are forwarded to the swap module to identify the training samples associated with noisy labels. Furthermore, the ranking error values are forwarded to the flipping module for the correction process.

\noindent\textbf{Flipping Module:}
The flipping module aims at flipping the noisy labels identified by both networks. To this end, the flipping module consists of two components: i) noisy class selector (NCS); and ii) noisy class flipper (NCF). The NCS takes the previously calculated ranking errors as input. The NCS component first compares the predicted labels by both networks to identify the labels with higher uncertainty (i.e., labels associated with high ranking error). Then, it eliminates the predictions that are not similar between two networks (i.e., labels that are not noisy). When both networks agree on the noisy labels, the ranking errors associated with the noisy labels from both networks are summed up. Then, the NCS component selects the labels with the largest ranking errors to apply the flipping. After flipping, an additional group lasso component recalculates the ranking losses for the given mini-batch with the flipped labels. It is worth noting that the flipped labels are also used for calculating the final BCE loss.

\noindent\textbf{Swap Module:}
As shown in Fig. \ref{fig:mmd} the swap module is injected between the two collaborative networks for exchanging the ranking information between the networks. The swap module takes the BCE and the ranking loss functions into consideration and eliminates the detected noisy samples from back-propagation. In detail, the swap module adds the ranking losses to the BCE losses. The swapping ranking losses are defined as follows:

\begin{equation} \label{eq:3.13}
\begin{split}
\mathit{B_{i}^{f} =  \mathit{L}_{f} (\mathbf{x}_{i}) + \gamma \mathit{Lasso_{f}(\mathbf{x}_{i})} } \ \ ,  \ \ {\mathbf{x_{i}} \in \mathbf{X}_{b} }  \;
\\
\mathit{B_{i}^{g} =  \mathit{L}_{g} (\mathbf{x}_{i}) + \gamma \mathit{Lasso_{g}(\mathbf{x}_{i})} } \ \ ,  \ \ {\mathbf{x_{i}} \in \mathbf{X}_{b} }  \;
\end{split}
\end{equation}
\noindent where $\mathit{B_{i}^{f}}$ and $\mathit{B_{i}^{g}}$ are the swapping ranking losses for the networks $\mathit{f}$ and $\mathit{g}$, respectively. $\mathbf{X}_{b}$ is the set of training images in the given mini-batch $b$. $\gamma$ is the trade off parameter representing the strength of $\mathit{Lasso_{f}}$ and $\mathit{Lasso_{g}}$. 
After calculating the swapping ranking losses, $\mathit{R}$ samples from the given mini-batch that are associated with the lowest swapping ranking loss values for each network are selected, and the ranking information between the two networks is exchanged. In this way, the network $\mathit{f}$ uses $\mathit{R}$ samples associated with the lowest loss values identified by the network $\mathit{g}$ to update its weights by minimizing $\mathit{Loss}_{f}^{b}$ in (\ref{eq:3.3}). Similarly, network $\mathit{g}$ uses the identified $\mathit{R}$ samples associated with the lowest loss values by $\mathit{f}$ to update its weights by minimizing $\mathit{Loss}_{g}^{b}$ in (\ref{eq:3.3}).


Through the end-to-end training of the two collaborative networks the parameters of the networks $f$ and $g$ are learned by minimizing the $\mathit{Loss}_{f}^{b}$ and $\mathit{Loss}_{g}^{b}$ losses introduced in (\ref{eq:3.3}). After the training phase, both networks are used to classify each image in the archive.

\section{experimental results}
\label{sec:experiment}
We conducted experiments on the Ireland subset of the BigEarthNet (denoted as IR-BigEarthNet) benchmark archive \cite{sumbul2019bigearthnet}, which consists of $15,894$ Sentinel-2 multispectral images acquired between June 2017 and May 2018 over Ireland. Each image is annotated by multiple land-cover classes provided by the CORINE Land Cover Map (CLC) database of the year 2018. We employed the BigEarthNet-19 land-cover class nomenclature proposed in~\cite{sumbul2021bigearthnetmm} in the experiments. In addition, we eliminated $5$ classes represented with a very small number of training images in the dataset, resulting in $12$ classes in total. The number of labels associated with each image varies between $1$ and $7$.

We employed the ResNet\cite{he2016deep} and DenseNet \cite{densenet} models as baselines to evaluate the proposed CCML method. The same architectures were considered as backbones for our CCML (for collaborative networks $f$ and $g$). We trained the models for 100 training epochs with the Adam optimizer using an initial learning rate of 0.001. We employed the same hyperparameters for the baseline models and the CCML. Within the swap module of the CCML, we used $75\%$ of the samples associated with small loss values at each iteration for swapping. It is crucial to choose the right time to start flipping potentially noisy labels. This is because of the fact that the networks may not be stable in the early stage of the training, resulting in wrong predictions. To address this problem, we activated the flipping module after reaching $90\%$ of epochs. The flipping module flips only the $5\%$ of the agreed classes at each iteration. To inject synthetic label noise in our experiments, we first randomly selected $n\%$ of samples from each mini-batch. Then, we randomly flipped $n\%$ of the labels from the selected samples. The value of $n$ was varied from $20$ to $50$ with a step size increment of $10$. The results were provided in precision, recall and $F_1$ score with the micro averaging strategy. The reader is referred to \cite{durand2019learning} for details on the considered metrics.

\begin{table}[htb]
\centering
\renewcommand{\arraystretch}{1.4}
\setlength{\tabcolsep}{5.8pt}
\captionsetup{justification=justified,singlelinecheck=false}
\caption{Precision, Recall and $F_1$ scores obtained by the proposed CCML and the baseline architecture (ResNet) \cite{he2016deep} under different noise rates.}
\label{tab:ire12det}
\begin{adjustbox}{width=.48\textwidth,center}
\begin{tabular}{|c|c|c|c|c|c|c|}
\cline{1-7}
Injected& \multicolumn{2}{c|}{Precision (\%)} & \multicolumn{2}{c|}{Recall (\%)} & \multicolumn{2}{c|}{$F_1$ (\%)} \\ \cline{2-7} 
Noise & Baseline & Proposed & Baseline & Proposed & Baseline & Proposed \\ 
Rate & (ResNet) & CCML & (ResNet) & CCML & (ResNet) & CCML \\ \hline 
20\% & 87.8 & \textbf{90.2} & \textbf{68.7} & \textbf{68.7} & 77.1 & \textbf{78.0} \\ \cline{1-7} 
30\% & 84.0 & \textbf{88.2} & 67.2 & \textbf{68.9} & 74.7 & \textbf{77.4} \\ \cline{1-7} 
40\% & 76.4 & \textbf{88.4} & 65.1 & \textbf{69.3} & 70.3 & \textbf{77.7} \\ \cline{1-7} 
50\% & 62.5 & \textbf{87.5} & 57.6 & \textbf{62.1} & 60.0 & \textbf{72.6} \\ \hline
\end{tabular}
\end{adjustbox}
\end{table}

\begin{table}[htb]
\centering
\renewcommand{\arraystretch}{1.4}
\setlength{\tabcolsep}{4pt}
\captionsetup{justification=justified,singlelinecheck=false}
\caption{Precision, Recall and $F_1$ scores obtained by the proposed CCML and the baseline architecture (DenseNet) \cite{densenet} under different noise rates.}
\label{tab:ire12dense}
\begin{adjustbox}{width=.48\textwidth,center}
\begin{tabular}{|c|c|c|c|c|c|c|}
\cline{1-7}
Injected& \multicolumn{2}{c|}{Precision (\%)} & \multicolumn{2}{c|}{Recall (\%)} & \multicolumn{2}{c|}{$F_1$ (\%)} \\ \cline{2-7} 
Noise & Baseline & Proposed & Baseline & Proposed & Baseline & Proposed \\ 
Rate & (DenseNet) & CCML & (DenseNet) & CCML & (DenseNet) & CCML \\ \hline
20\% & 89.2 & \textbf{89.6} & 68.4 & \textbf{77.4} & 77.1 & \textbf{78.1} \\ \cline{1-7} 
30\% & 91.8 & \textbf{92.0} & 66.2 & \textbf{66.7} & 76.9 & \textbf{77.3} \\ \cline{1-7} 
40\% & 85.6 & \textbf{89.0} & 68.5 & \textbf{68.7} & 76.1 & \textbf{77.5} \\ \cline{1-7} 
50\% & 55.3 & \textbf{85.1} & 62.5 & \textbf{66.4} & 58.7 & \textbf{74.6} \\ \hline
\end{tabular}
\end{adjustbox}
\end{table}

\begin{figure*}
\centering
\renewcommand{\arraystretch}{2}
\setlength{\tabcolsep}{-2.1pt}
\centering
\begin{tabular}{cccc}
\hspace{-10pt}
    	\begin{subfigure}{0.28\linewidth}
    	\centering
    	\captionsetup{justification=centering, margin={1cm,0cm}}
    	\begin{tikzpicture}[scale = 0.90]%
    		\begin{axis}[
            legend columns=1,
            ylabel near ticks,
            legend style={font=\scriptsize},
            xtick={0,10,20,...,50},
            width=\linewidth,
       		height=\linewidth,
    		legend pos=south west,
            ymin=0.1,
            xlabel= Injected Noise Rate (\%),
            ylabel= $F_1$ Score,
            xmin=20,xmax=50,
            ymin=0.75,ymax=1.0]
            \addplot+[name path=reused,mark=*,color=blue,mark options={fill=white},line width=0.8pt] table [x=perc, y=c3, col sep=comma] {data/base.csv};
            \addplot+[name path=capacity,color=red,mark=*,mark options={fill=white},line width=0.8pt] table [x=perc, y=c3, col sep=comma] {data/ccml.csv};
    		\legend{Baseline (ResNet), Proposed CCML}
    		\end{axis}
    	\end{tikzpicture}
    \end{subfigure}
    \hspace{-10pt}
    & 
    \begin{subfigure}{0.28\linewidth}
    
    \centering
    	\captionsetup{justification=centering, margin={1cm,0cm}}
    	\hspace{-10pt}
    	\begin{tikzpicture}[scale = 0.90]%
    		\begin{axis}[
            legend columns=1,
            legend style={font=\scriptsize},
            ylabel near ticks,
            xtick={0,10,20,...,50},
            width=\linewidth,
       		height=\linewidth,
    		legend pos=south west,
            ymin=0.1,
            xlabel= Injected Noise Rate (\%),
            ylabel= $F_1$ Score,
            xmin=20,xmax=50,
            ymin=0.75,ymax=1.0]
            \addplot+[name path=reused,mark=*,color=blue,mark options={fill=white},line width=0.8pt] table [x=perc, y=c12, col sep=comma] {data/base.csv};
            \addplot+[name path=capacity,color=red,mark=*,mark options={fill=white},line width=0.8pt] table [x=perc, y=c12, col sep=comma] {data/ccml.csv};
    		\legend{Baseline (ResNet), Proposed CCML}
    		\end{axis}
    	\end{tikzpicture}
    \end{subfigure}
    \hspace{-10pt}
    &
    \begin{subfigure}{0.28\linewidth}
    \centering
    	\captionsetup{justification=centering, margin={1cm,0cm}}
    	\hspace{-10pt}
    	\begin{tikzpicture}[scale = 0.90]%
    		\begin{axis}[
            legend columns=1,
            legend style={font=\scriptsize},
            ylabel near ticks,
            xtick={0,10,20,...,50},
            width=\linewidth,
       		height=\linewidth,
    		legend pos=south west,
            ymin=0.1,
            xlabel= Injected Noise Rate (\%),
            ylabel= $F_1$ Score,
            xmin=20,xmax=50,
            ymin=-0.09,ymax=0.80]
            \addplot+[name path=reused,mark=*,color=blue,mark options={fill=white},line width=0.8pt] table [x=perc, y=c11, col sep=comma] {data/base.csv};
            \addplot+[name path=capacity,color=red,mark=*,mark options={fill=white},line width=0.8pt] table [x=perc, y=c11, col sep=comma] {data/ccml.csv};
    		\legend{Baseline (ResNet), Proposed CCML}
    		\end{axis}
    	\end{tikzpicture}
    \end{subfigure}
    \hspace{-10pt}
    &
    \begin{subfigure}{0.28\linewidth}
        	\centering
    	\captionsetup{justification=centering, margin={1cm,0cm}}
    	\hspace{-10pt}
    	\begin{tikzpicture}[scale = 0.90]%
    		\begin{axis}[
            legend columns=1,
            legend style={font=\scriptsize},
            ylabel near ticks,
            xtick={0,10,20,...,50},
            ytick={0,0.1,0.2,...,0.7},
            width=\linewidth,
       		height=\linewidth,
    		legend pos=south west,
            ymin=0.1,
            xlabel= Injected Noise Rate (\%),
            ylabel= $F_1$ Score,
            xmin=20,xmax=50,
            ymin=0.35,ymax=0.6]
            \addplot+[name path=reused,mark=*,color=blue,mark options={fill=white},line width=0.8pt] table [x=perc, y=c9, col sep=comma] {data/base.csv};
            \addplot+[name path=capacity,color=red,mark=*,mark options={fill=white},line width=0.8pt] table [x=perc, y=c9, col sep=comma] {data/ccml.csv};
    		\legend{Baseline (ResNet), Proposed CCML}
    		\end{axis}
    	\end{tikzpicture}
    \end{subfigure}
    \hspace{-10pt}\\
    \hspace{30pt}(a)& \hspace{20pt}(b) & \hspace{20pt}(c) & \hspace{20pt}(d)
    \vspace{-8pt}
    \end{tabular}
    \captionsetup{justification=justified,singlelinecheck=false}
    \caption{Different noise rates versus class based $F_1$ scores obtained by the ResNet baseline and the proposed CCML for four selected classes: (a) Pastures; (b) Marine waters; (c) Inland wetlands; and (d) Moors, heathland and sclerophyllous vegetation.}
    \label{fig:ireclass}
\end{figure*}
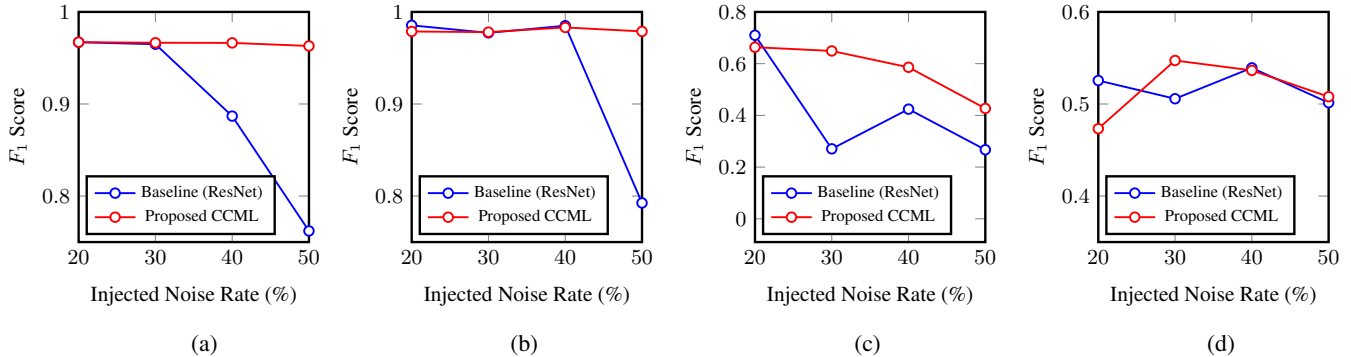

The results obtained when the baseline architectures are ResNet and DenseNet are presented in Tables \ref{tab:ire12det} and \ref{tab:ire12dense}, respectively. From the tables, one can observe that the proposed CCML improves the multi-label classification performance of the baseline architectures under all metrics. As an example, in Table \ref{tab:ire12det} when the baseline architecture is ResNet and the injected noise rate is $50\%$, the proposed CCML (with ResNet backbone) obtained $25\%$ and $10\%$ higher precision and $F_1$ score, respectively. From Table \ref{tab:ire12dense}, when the baseline architecture is DenseNet, under the synthetically added noise rate of $50\%$, the proposed CCML (with DenseNet backbone) sharply outperforms the baseline model by $30\%$ and $15\%$ in precision and $F_1$ score, respectively. These results also prove that the proposed CCML has the capability of reducing the negative effect of label noise independently from the selected backbone architecture. We also evaluate our proposed CCML under lower (synthetically added) noise rates. From this analysis, we observe that the CCML does not significantly improve the multi-label classification performance. This is mainly because of the fact that $25\%$ of the training samples is excluded when CCML is applied. For space constraints, we could not report these results. 

We also analyzed the class-based performances of the proposed CCML and the baseline models. Fig. \ref{fig:ireclass} shows the class-based $F_1$ scores associated with four land-cover classes present in IR-BigEarthNet when the baseline model as well as the backbone of the CCML were selected as ResNet. From the results, one can observe that the baseline model and the proposed CCML can accurately learn the classes represented by a high number of training images. As an example, the ``Marine waters'' class is associated with a high number of training images and the $F_1$ scores of this class are high under different noise rates for the baseline and the proposed CCML. By analyzing the results in Fig. \ref{fig:ireclass}, one can also see that the baseline and the CCML performances are comparable over lower rates of label noise ($20\%$ and $30\%$) in some classes (e.g., ``Marine waters'' and "Pastures"). However, the proposed CCML maintains relatively high performance under high noise rates ($40\%$ and $50\%$) compared to the baseline model, which demonstrates the stability of the proposed CCML under high noise rates. Furthermore, the classes with a small number of training images (e.g., ``Moors, heathland and sclerophyllous vegetation'') are more affected by the label noise, and classification performance is low for both methods (see Fig. \ref{fig:ireclass}d). 

We also analyzed the effect of each module in CCML. From the results, we observed that the discrepancy and swap modules are fundamental for the success of the CCML. Without the discrepancy module, CCML fails to learn diverse representations of the data (which leads to the fact that the two parallel networks learn similar features, contradicting the main idea of CCML). CCML exploits the group lasso module along with the cross-entropy to detect noisy labels. The exclusion of the group lasso degrades the performance since CCML fails to identify different types of noise individually. However, removing the flipping module does not decrease the performance substantially because of the delayed initiation of the flipping module. For space constraints, we could not report these results.



\section{Conclusion}
\label{sec:conclusion}
In this paper, a novel Consensual Collaborative Multi-label Learning (CCML) method has been presented. The proposed method overcomes the adverse effects of multi-label noise for the classification of RS image scenes. To this end, it includes four main modules: 1) discrepancy module; 2) group lasso module; 3) flipping module; and 4) swap module. Unlike the existing methods in RS, the proposed CCML automatically identifies two different types of multi-label noise (i.e., missing and wrong class label annotations) without making any prior assumptions. In addition, CCML is an architecture-independent method and thus it is suitable to be used with different network architectures.

We would like to note that the CCML requires to exclude a portion of the training samples during the training phase. This approach may reduce classification accuracy when the amount of label noise in the training set is very small. To avoid this issue, as a future work, we plan to investigate an adaptive ranking loss function to adjust the amount of sample removal according to the estimated noise rate. It is also worth noting that the CCML assumes that each class is represented by a sufficient number of training images. However, the availability of such training sets may not always be possible in the operational RS MLC applications. As a future work, to address this issue, we plan to develop an adaptive class-weighted loss function and to investigate data augmentation techniques.

\section{Acknowledgement}
This work is supported by the German Ministry for Education and Research as BIFOLD - Berlin Institute for the Foundations of Learning and Data (01IS18025A).

\bibliographystyle{IEEEbib}
\bibliography{refs}

\end{document}